\documentclass[10pt,twocolumn,letterpaper]{article}

\usepackage{cvpr}
\usepackage{times}
\usepackage{epsfig}
\usepackage{graphicx}
\usepackage{amsmath}
\usepackage{amssymb}

\usepackage{subcaption}
\usepackage{multirow}
\usepackage[labelfont=bf]{caption}
\graphicspath{{./figures/}}
\newcommand{\architecture}[1]{\emph{#1}}
\newcommand{\arch}[1]{\emph{#1}}

\newcommand{\mrcell}[1]{\multicolumn{1}{c|}{\begin{tabular}[c]{@{}c@{}} #1 \end{tabular}}}
\newcommand{\lmrcell}[1]{\multicolumn{1}{|c|}{\begin{tabular}[c]{@{}c@{}} #1 \end{tabular}}}

\usepackage[pagebackref=true,breaklinks=true,letterpaper=true,colorlinks,bookmarks=false]{hyperref}

\cvprfinalcopy 

\def\cvprPaperID{2863} 

\ifcvprfinal\fi
\begin{document}

\title{Training Competitive Binary Neural Networks from Scratch}

\author{Joseph Bethge\footnotemark[1] , Marvin Bornstein\footnotemark[2] , Adrian Loy\footnotemark[2] , Haojin Yang\footnotemark[1] , Christoph Meinel\footnotemark[1]  \\
Hasso Plattner Institute, University of Potsdam, Germany\\
P.O. Box 900460, Potsdam D-14480\\
\footnotemark[1] {\tt\small firstname.surname@hpi.de}, \footnotemark[2] {\tt\small firstname.surname@student.hpi.de}
}

\maketitle

\begin{abstract}

Convolutional neural networks have achieved astonishing results in different application areas.
Various methods that allow us to use these models on mobile and embedded devices have been proposed.
Especially binary neural networks are a promising approach for devices with low computational power.
However, training accurate binary models from scratch remains a challenge.
Previous work often uses prior knowledge from full-precision models and complex training strategies.
In our work, we focus on increasing the performance of binary neural networks without such prior knowledge and a much simpler training strategy.
In our experiments we show that we are able to achieve state-of-the-art results on standard benchmark datasets.
Further, to the best of our knowledge, we are the first to successfully adopt a network architecture with dense connections for binary networks, which lets us improve the state-of-the-art even further.
Our source code can be found online: \\
\url{https://github.com/hpi-xnor/BMXNet-v2}

\end{abstract}


\section{Introduction}
\label{sec:intro}

Nowadays, significant progress through research is made towards automating different tasks of our everyday lives.
From vacuum robots in our homes to entire production facilities run by robots, many tasks in our world are already highly automated.
Other advances, such as self-driving cars, are currently being developed and depend on strong machine learning solutions.
The amount of apps on smartphones, which adopt deep learning techniques to solve a variety of tasks, is rising rapidly and will likely continue to do so in the future.
All these devices have limited computational power, often while trying to minimize energy consumption, but have many use cases for machine learning. 

We will consider the example of a fully automated self-driving car. 
It is crucial for such a system to achieve high accuracy coupled with guaranteed real-time image processing.
Furthermore, the image processing system needs to be hosted in the car itself, as a stable Internet connection with low latency cannot be guaranteed in this setting.
This requirement limits the available computational power and memory, but at the same time profits from a low energy consumption.
A promising technique that can deal well with these conditions are Binary Neural Networks (BNNs).
In a BNN the commonly used full-precision weights of a convolutional neural network are replaced with binary weights.
This results in a storage compression by a factor of 32$\times$ and allows for significantly more efficient inference on CPU-only architectures.


We discuss existing approaches in \autoref{sec:related}.
Moreover, we identified three ways to increase the accuracy of a binary model and describe how we applied them to a binary network with dense shortcut connections:
removing bottleneck designs, increasing the number of shortcut connections throughout the network, and replacing certain layers with full-precision layers.
We describe these and other common techniques together with our implementation details in \autoref{sec:method}.
Afterwards, we discuss the results of our approach on the MNIST, CIFAR10 and ImageNet datasets in \autoref{sec:experiments}.
We evaluate the influence of the previously described techniques on existing approaches and with our proposed model based on dense shortcut connections.
The results show that we can reach state-of-the-art results for existing architectures and improve results even further with our proposed model.
Finally, we examine future ideas and conclude our work in \autoref{sec:conclusion}.

Summarized, our contributions in this paper are:
\begin{itemize}
\itemsep0em 
    \item We present a simple training strategy for binary models without using a pretrained full-precision model.
    \item We provide empirical evidence that this strategy does not benefit from other commonly used methods, \eg, scaling factors or usage of custom gradient calculation.
    \item We show that increasing the number of shortcut connections improves the classification accuracy of BNNs significantly and show a novel way to create efficient binary models based on dense shortcut connections.
    \item We reach state-of-the-art accuracy compared to other approaches for different model architectures and sizes.
\end{itemize}

\section{Related Work}
\label{sec:related}



In this section we present related work for binarization and compression techniques.


There are two main approaches which allow for execution on mobile devices by accelerating inference:
On the one hand, information in a CNN can be compressed through compact network design.
These designs use full-precision floating point numbers as weights, but reduce the total number of parameters and operations through clever network design, while preventing loss of accuracy.
On the other hand, information can be compressed by avoiding the common usage of full-precision floating point weights, which use 32 bits of storage.
Instead, quantized floating-point numbers with lower precision (\eg 4 bit of storage) or even binary (1 bit of storage) weights are used in these approaches.


First, we present a selection of techniques which utilize the former method.
The first of these approaches, \emph{SqueezeNet}, was presented by Iandola \etal \cite{Iandola2016}.
The authors replace a large portion of 3$\times$3 filters with smaller 1$\times$1 filters in convolutional layers and reduce the number of input channels to the remaining 3$\times$3 filters for a reduced number of parameters.
Additionally, they facilitate late downsampling to maximize their accuracy and use \emph{deep compression} \cite{Han2015} for an overall model size of 0.5 MB.

A different approach, \emph{MobileNets}, was implemented by Howard \etal \cite{Howard2017}.
They use a depth-wise separable convolution where convolutions apply a single 3$\times$3 filter to each input channel.
Subsequently, a 1$\times$1 convolution is applied to combine their outputs.
Zhang \etal \cite{Zhang2017} use channel shuffling to achieve group convolutions in addition to depth-wise convolution.
Their \emph{ShuffleNet} achieves comparably lower error rate for the same number of operations needed for \emph{MobileNets}.
These approaches reduce memory requirements, but still require GPU hardware for efficient training and inference.
A strategy to accelerate the computation of all these methods for CPUs has yet to be developed.


In contrast to this, approaches which use binary weights instead of full-precision weights achieve compression and acceleration.
However, the drawback usually is a severe drop in accuracy.
These approaches are based on \emph{Binarized Neural Networks}, introduced by Hubara \etal \cite{Courbariaux2016}, where weights and activations are restricted to +1 and -1.
They provide efficient calculation methods for the equivalent of a matrix multiplication by using $\mathrm{xnor}$ and $\mathrm{popcount}$ operations.
\emph{XNOR-Nets}, published by Rastegari \etal \cite{Rastegari2016}, improved the performance of binary neural networks by introducing changes to the network layout.
Furthermore, they include a channel-wise scaling factor to reduce the approximation error of full-precision weights.
Another approach, called \emph{DoReFa-Net}, was presented by Zhou \etal \cite{Zhou2016}.
They focus on quantizing the gradients together with different bit-widths (down to binary values) for weights and activations and replace the channel-wise scaling factor with one constant scalar for all filters.
A different attempt to strictly use nothing except binary weights is taken in \emph{ABC-Nets} by Lin \etal \cite{lin2017towards}.
They use 3 to 5 binary weight bases to approximate full-precision weights.
This approximation increases model complexity and size, but reduces the gap between the accuracy of full-precision and binary networks to 5\%.
Wan \etal \cite{Wan_2018_ECCV} improved accuracy by using binary weights and ternary activations in their \emph{Ternary-Binary Network}. 
They train their model from scratch, but they have more operations compared to fully binary models (without an increase in memory consumption).
In \emph{Bi-Real Net}, Liu \etal \cite{Liu_2018_ECCV} modify the \arch{ResNet} architecture by adding additional shortcuts and reducing the size of the convolution layers.
They propose a change of gradient computation during backpropagation compared to other approaches.
\architecture{Bi-Real Nets} are trained using a complex training strategy to fine-tune a pretrained full-precision network to create a binary model with 56.4\% accuracy.
Our work differs from their approach, as we directly train a binary network from scratch.




\section{Methodology}
\label{sec:method}


In this section we first provide the major implementation principles of the framework we use for implementing and training binary models.
Following this, we examine the usage of scaling factors.
Finally, we discuss design principles for binary network layouts and introduce a novel binary model architecture based on \arch{DenseNets}.

\subsection{Implementation of Binary Layers}
\label{sec:implementation-details}

Our implementation is based on the BMXNet framework first presented by Yang \etal \cite{HPI_xnor}, which itself is based on the MXNet framework.
We use the sign function for activation, thus transforming floating-point values into binary values:
\begin{equation}
    \mathrm{sign}(x) = \begin{cases} 
    +1 ~\text{if}~ x \geq 0, \\
    -1 ~\text{otherwise}.
    \end{cases}
\end{equation}
The implementation uses a Straight-Through Estimator (STE) \cite{hinton2012neural} with the addition, that it cancels the gradients, when the inputs get too large, as proposed by Hubara \etal \cite{Courbariaux2016}.
The gradient canceling helps the optimization process, since backpropagation no longer increases the absolute value of an input larger than the clipping threshold (which has no actual effect on the loss because $\mathrm{sign}$ does not depend on the absolute value).
Let $c$ denote the objective function, $r_i$ be a real number input, and $r_o\in\{-1,+1\}$ a binary output.
Furthermore, $t_\mathrm{clip}$ is the threshold for clipping gradients, which was set to $t_\mathrm{clip}=1$ in previous works~\cite{Courbariaux2016,Zhou2016}.
Then, the resulting STE is:
\begin{align}
    \text{Forward:}& ~r_o=\mathrm{sign}(r_i)~. \\
    \text{Backward:}& ~\frac{\partial c}{\partial r_i}=\frac{\partial c}{\partial r_o}1_{|r_i|\leq t_\mathrm{clip}}~.
\end{align}
Liu \etal \cite{Liu_2018_ECCV} claim that a tighter approximation, called $\mathrm{approxsign}$, can be made by replacing the backward pass with
\begin{align}
    ~\frac{\partial c}{\partial r_i}=\frac{\partial c}{\partial r_o}1_{|r_i|\leq t_\mathrm{clip}}\cdot\begin{cases} 
    2-2r_i ~\text{if}~ r_i \geq 0, \\
    2+2r_i ~\text{otherwise}.
    \end{cases}
\end{align}
Since this could also benefit when training a binary network from scratch, we evaluated this in our experiments.

A large amount of calculations in full-precision networks is usually spent on calculating dot products of matrices, as needed for fully connected and convolutional layers.
The computational cost of binary neural networks can be highly reduced by using the $\mathrm{xnor}$ and $\mathrm{popcount}$ CPU instructions, first presented by Rastegari \etal \cite{Rastegari2016}.
They show that the matrix multiplication of a binary input $x$ and weight $w$ can be replaced as follows ($n$ is the number of weights):
\begin{equation}
\label{eqn:xnor-popcount}
    x \cdot w = 2 \odot \mathrm{bitcount}(\mathrm{xnor}(x',w')) - n~.
\end{equation}
Note, that $x'$ and $w'$ are converted from $x$ and $w$ by replacing $\{-1,+1\}$ with $\{0,1\}$.
This means normal training methods with GPU acceleration (\eg cuDNN implementation) can be used (the left side of \autoref{eqn:xnor-popcount}).
Afterwards, we can take advantage of the fast CPU implementation with $\mathrm{xnor}$ and $\mathrm{popcount}$ (the right side of \autoref{eqn:xnor-popcount}) without any accuracy loss (an example can be found in the supplementary material).
To further speedup the final CPU implementation the adjustment by the number of weights can be learned during training (derived from \autoref{eqn:xnor-popcount}):
\begin{equation}
\label{eqn:xnor-popcount-swapped}
    \frac{x \cdot w + n}{2} = \mathrm{bitcount}(\mathrm{xnor}(x',w'))~.
\end{equation}
Further, we decide to use no weight decay during training.
This was done in previous work before without much explanation \cite{Liu_2018_ECCV}, so we add our rationale here:
Since gradient canceling already prevents the network from optimizing to absolute values larger than the clipping threshold (i.e.~the values are already optimal for the current mini-batch), adding weight decay would move these weights away from their optimal values.

\subsection{Scaling Methods}
\label{sec:scaling}



\begin{figure}
    \centering
    \includegraphics[width=0.8\linewidth]{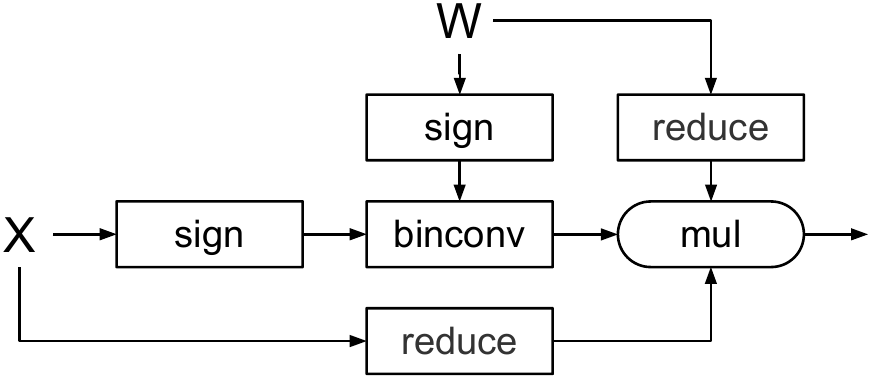}
    \caption{General computation graph for a scaled binary convolution. \emph{binconv} is a convolution based on the optimized operation in \autoref{eqn:xnor-popcount}. \emph{reduce} computes a scaling factor for activations or weights (can be chosen differently), and \emph{mul} is a multiplication operation.}
    \label{fig:scaling}
\end{figure}

In this section, we discuss the usage of a scaling factor during training.
Binarization will always introduce an approximation error compared to a full-precision signal.
In their analysis, Zhou \etal \cite{Zhou2017} show that this error linearly degrades the accuracy of a CNN.
One way to reduce the approximation error, is to use scaling factors \cite{Zhou2016,Rastegari2016,Liu_2018_ECCV}.
Generally, they follow the structure as in \autoref{fig:scaling}. 


Rastegari \etal \cite{Rastegari2016} choose $\mathrm{reduce}(w) = f_{s_w}(w) = \frac{1}{n}||w||_{1,1}$ for each weight filter $w$.
They further propose an efficient method for scaling each feature (i.e.~$\mathrm{reduce}(x) = \mathbf{K}$ referring to their paper \cite{Rastegari2016}).

In contrast, Zhou \etal \cite{Zhou2016} reported that a filter-wise weight scaling does not yield improvements.
They use one scalar for all weight filters instead, allowing them to also use a binary convolution in the backward pass.
Liu \etal \cite{Liu_2018_ECCV} suggest to use the weight scaling $f_{s_w}$ only in the backward pass to achieve \emph{magnitude aware gradients}.

The scaling factors should help binary convolutions to increase the value range.
Producing results closer to those of full-precision convolutions and reducing the approximation error.
However, these different scaling values influence specific output channels of the convolution.
Therefore, a BatchNorm \cite{ioffe2015batch} layer directly after the convolution (which is used in \architecture{ResNet} and \architecture{DenseNet} architectures) theoretically minimizes the difference between a binary convolution with scaling and one without.

Thus, we hypothesize that learning a useful scaling factor is made inherently difficult by BatchNorm layers.
We empirically evaluated this in our experiments (see \autoref{sec:results-scaling}), but want to note that this reasoning might not apply if a binary model is fine-tuned from a full-precision model. 

\subsection{Network Architectures}
\label{sec:architectures}

\begin{figure}[t]
\captionsetup[subfigure]{justification=centering}
\begin{center}
\begin{subfigure}[t]{0.32\linewidth}
   \centering
   \includegraphics[width=0.915\linewidth]{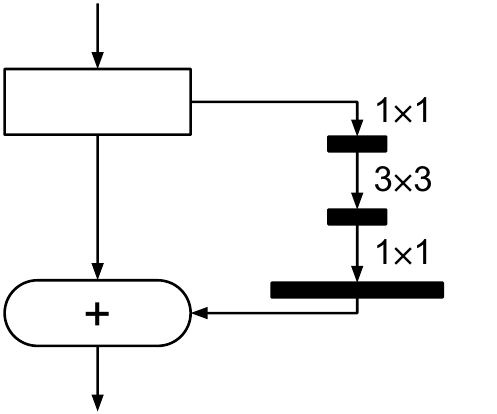}
   \caption{ResNet \\ (bottleneck)}
   \label{fig:netblocks-resnet-bottleneck}
\end{subfigure}
\begin{subfigure}[t]{0.32\linewidth}
   \centering
   \includegraphics[width=0.915\linewidth]{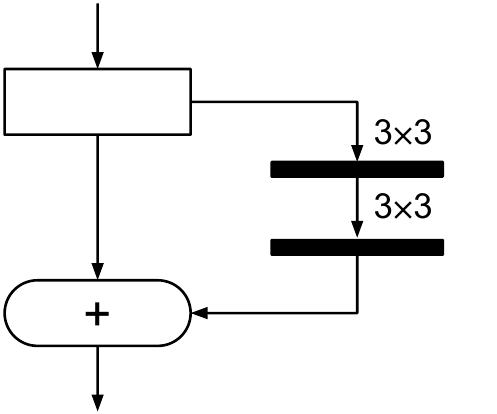}
   \caption{ResNet \\ (no bottleneck)}
   \label{fig:netblocks-resnet}
\end{subfigure}
\begin{subfigure}[t]{0.32\linewidth}
   \centering
   \includegraphics[width=0.915\linewidth]{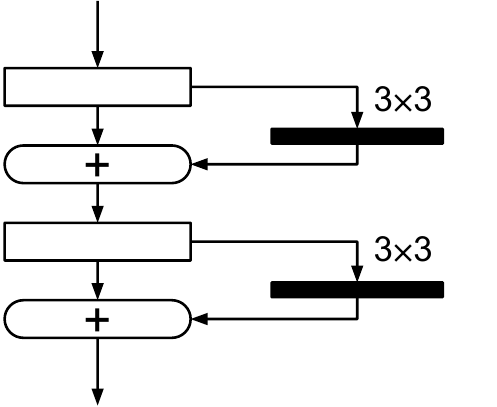}
   \caption{ResNetE \\ (added shortcut)}
   \label{fig:netblocks-resnete}
\end{subfigure}
\begin{subfigure}[t]{0.32\linewidth}
   \centering
   \includegraphics[width=0.99\linewidth]{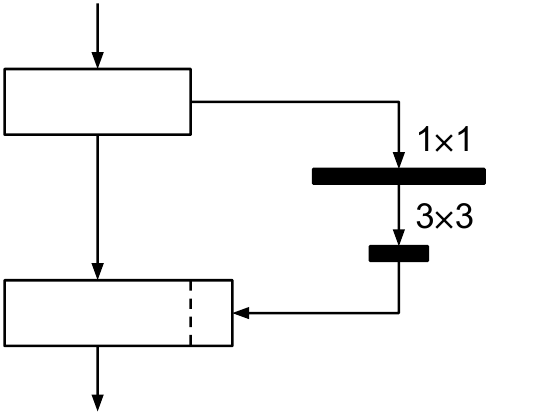}
   \caption{DenseNet \\ (bottleneck)}
   \label{fig:netblocks-densenet-bottleneck}
\end{subfigure}
\begin{subfigure}[t]{0.32\linewidth}
   \centering
   \includegraphics[width=0.99\linewidth]{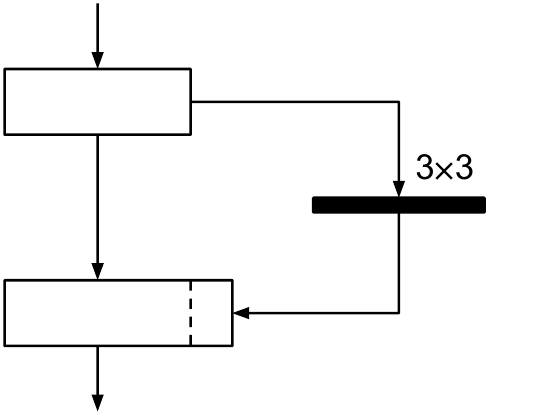}
   \caption{DenseNet \\ (no bottleneck)}
   \label{fig:netblocks-densenet}
\end{subfigure}
\begin{subfigure}[t]{0.32\linewidth}
   \centering
   \includegraphics[width=0.99\linewidth]{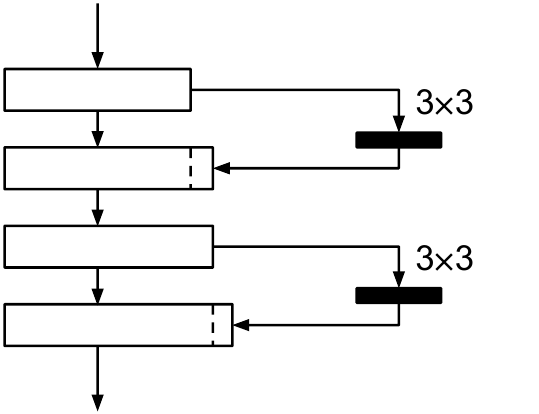}
   \caption{DenseNetE \\ (our suggestion)}
   \label{fig:netblocks-densenet-split}
\end{subfigure}
\end{center}
\caption{
   A single building block of different network architectures (the length of bold black lines represents the number of filters).
    (\subref{fig:netblocks-resnet-bottleneck}) The original \architecture{ResNet} design features a bottleneck architecture.
    A low number of filters reduces information capacity for binary neural networks.
    (\subref{fig:netblocks-resnet}) A variation of the \architecture{ResNet} architecture without the bottleneck design.
    The number of filters is increased, but with only two convolutions instead of three.
    (\subref{fig:netblocks-resnete}) The \architecture{ResNet} architecture with an additional shortcut, first introduced by Liu \etal \cite{Liu_2018_ECCV}.
    (\subref{fig:netblocks-densenet-bottleneck}) The original \architecture{DenseNet} design with a bottleneck in the second convolution operation.
    (\subref{fig:netblocks-densenet}) The \architecture{DenseNet} design without a bottleneck.
    The two convolution operations are replaced by one $3\times3$ convolution.
    (\subref{fig:netblocks-densenet-split}) Our suggested change to a \architecture{DenseNet} where a convolution with N filters is replaced by two layers with $\frac{N}{2}$ filters each.
}
\label{fig:netblocks}
\end{figure}
%


In this section we describe general concepts for binary deep neural network architectures first.
Afterwards, we show details about \architecture{ResNet} \cite{He2017} and our suggested binary \architecture{DenseNet} architecture \cite{Huang2016}.

Before thinking about model architectures, we must consider the main drawbacks of binary neural networks.
First of all, the information density is theoretically 32 times lower, compared to full-precision networks.
Research suggests, that the difference between 32 bits and 8 bits seems to be minimal and 8-bit networks can achieve almost identical accuracy as full-precision networks \cite{Han2015}.
However, when decreasing bit-width to four or even one bit (binary), the accuracy drops significantly \cite{Courbariaux2016,Zhou2016}.
Therefore, the precision loss needs to be alleviated through other techniques, for example by increasing information flow through the network.
We identified three main methods, which help to preserve information despite binarization of the model:

First, a binary model should use as many shortcut connections as possible in the network.
These connections allow layers later in the network to access information gained in earlier layers despite of information loss through binarization.
Such shortcut connections were proposed for full-precision model architectures in Residual Networks \cite{He2017} and Densely Connected Networks \cite{Huang2016}.
Furthermore, this means increasing the number of connections between layers should lead to better model performance, especially for binary networks.

Secondly, following the same idea, network architectures including bottlenecks are always a challenge to adopt.
A bottleneck architecture reduces the number of filters and values significantly between the layers, resulting in less information flow through binary neural networks.
Therefore we hypothesize, that either we need to eliminate the bottleneck parts or at least increase the number of filters in these bottleneck parts for binary neural networks to achieve best results.

The third way to preserve information (thus increasing model accuracy) comes from replacing certain crucial layers in a binary network with full precision layers.
The reasoning is as follows:
If layers are binarized, which do not have a shortcut connection, the information lost (due to binarization) can not be recovered in subsequent layers of the network.
This affects the first (convolutional) layer and the last layer (a fully connected layer which has a number of output neurons equal to the number of classes).
These layers generate the initial information for the network or consume the final information for the prediction, respectively.
Therefore, we use full-precision layers for the first and the final layer for all network architectures.
We follow authors of previous work on this decision \cite{Rastegari2016,Zhou2016}, who have empirically shown that binarizing these layers decreases accuracy by a large margin and that the saving of memory and operations is minimal.
Another crucial part of deep networks is the downsampling convolution which converts all previously collected information of the network to smaller feature maps with more channels (this convolution often has stride two and output channels equal to twice the number of input channels).
Any information lost in this downsampling process is effectively no longer available.
Therefore, it should always be considered whether these downsampling layers should be replaced with full-precision layers, even though it increases model size and number of operations.

In the following sections we show how all three methods are applied to a \arch{ResNet} (seen in previous work) and how we applied them to a \arch{DenseNet}. 

\begin{figure}[t]
\captionsetup[subfigure]{justification=centering}
\begin{center}
\begin{subfigure}[t]{0.57\linewidth}
   \centering
   \includegraphics[width=0.95\linewidth]{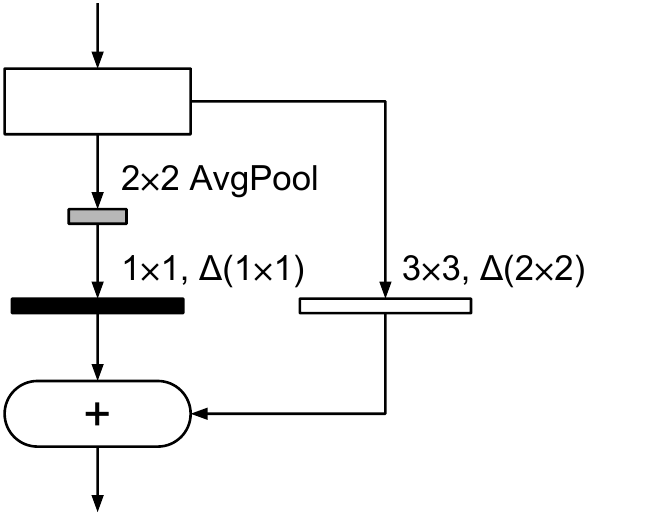}
   \caption{ResNet}
   \label{fig:transition-resnet}
\end{subfigure}
\begin{subfigure}[t]{0.39\linewidth}
   \centering
   \includegraphics[width=0.95\linewidth]{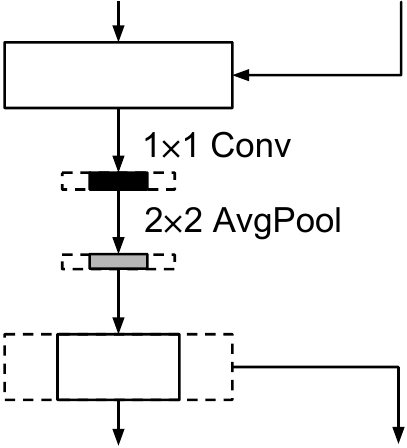}
   \caption{DenseNet}
   \label{fig:transition-densenet}
\end{subfigure}
\end{center}
\caption{
The downsampling layers of \arch{ResNet} and \arch{DenseNet}.
The bold black lines mark the downsampling layers which can be replaced with full-precision layers.
If we use this full-precision layer in a \arch{DenseNet}, we increase the reduction rate to reduce the number of channels (the dashed lines depict the number of channels without reduction).
}
\label{fig:downsampling}
\end{figure}

\subsubsection{ResNet Architecture}
\label{sec:architectures-resnet}

The \architecture{ResNet} architecture, introduced by He \etal \cite{He2017}, was the first model architecture that allowed to train models with 18 or more (up to 152) layers.
\architecture{ResNet} models combine the information of all previous layers with shortcut connections.
This is done by adding the input of a block to its output with an identity connection. 
Consequently, these shortcut connections add no extra weights and very little computational cost, while leading to more meaningful gradients in deeper layers.
The bottleneck of a \arch{ResNet} can be removed by replacing the three convolution layers (kernel sizes 1, 3, 1) of a regular \arch{ResNet} block with two three by three convolution layers with a higher number of filters (see \autoref{fig:netblocks-resnet-bottleneck}, \subref{fig:netblocks-resnet}).

Increasing the number of connections can be done by reducing the block size from two convolutions per block to one convolution per block, as proposed by Liu \etal \cite{Liu_2018_ECCV}.
This leads to twice the amount of shortcuts, as there are as many shortcuts as blocks, if the amount of layers is kept the same (see \autoref{fig:netblocks-resnete}).
However, their method also incorporates other changes to the \arch{ResNet} architecture. 
Therefore we call this specific change in the block design \architecture{ResNetE} (short for Extra shortcut).

Their second change was replacing the downsampling convolution layer (see \autoref{fig:transition-resnet}).
This was first proposed by Rastegari \etal \cite{Rastegari2016}, but neither work quantifies the exact accuracy gain nor the impact on the model size of this design choice.

\subsubsection{DenseNet Architecture}
\label{sec:architectures-densenet}

\architecture{DenseNets}, proposed by Huang \etal \cite{Huang2016}, use shortcut connections that, contrary to \arch{ResNets}, concatenate the input of a block to its output (see \autoref{fig:netblocks-densenet-bottleneck}, \subref{fig:netblocks-resnet}).
Therefore, new information gained in one layer can be reused throughout the entire depth of the network.
To reduce the total model size, the original full-precision architecture includes a bottleneck design for each block and additionally reduces the number of channels in transition layers.
This effectively keeps the network at a significantly smaller total size, even though the concatenation adds new information into the network every layer.
The number of newly appended features is called growth rate ($k$) and Huang \etal \cite{Huang2016} use $k=32$.
The bottleneck of the \arch{DenseNet} architecture can be modified by replacing the two convolution layers (kernel sizes 1 and 3) with one $3 \times 3$ convolution (see \autoref{fig:netblocks-densenet-bottleneck}, \subref{fig:netblocks-densenet}). 

However, our experiments showed that reusing the full-precision \architecture{DenseNet} architecture for binary neural networks does not achieve satisfactory performance, even after this change.
There are different possibilities to increase the capacity of a binary \architecture{DenseNet} architecture.
The growth rate can be increased (\eg $k=64, k=128$), we can use a larger number of blocks, or a combination of both.
Both individual approaches add roughly the same amount of parameters to the network.
To keep the number of parameters equal for a given \architecture{DenseNet} we can halve the growth rate and double the number of blocks at the same time (see \autoref{fig:netblocks-densenet-split}) or vice versa.
We assume that in this case increasing the number of blocks should provide better results compared to increasing the growth rate. 
This assumption is derived from our second hypothesis: favoring an increased number of connections over simply adding weights.
Similar to a \arch{ResNet} we refer to this adjustment as the \architecture{DenseNetE} architecture.
However, we note that the actual number of layers and growth rate can be chosen rather freely and evaluate different configurations.

Finally, another characteristic difference of a \arch{DenseNet} compared to a \arch{ResNet} is that the downsampling layer reduces the number of channels \cite{Huang2016}.
Our experiments showed, that without adjusting the architecture in these downsampling layers, a binary \architecture{DenseNet} achieves results of less than 40\% accuracy on ImageNet.
To preserve information flow in these parts of the network we found two options:
On the one hand, we can use no reduction at all, or at least use a lower reduction rate (using a higher number of channels compared to a full-precision architecture).
Since the number of channels is initially low in the first downsampling layer (\eg 384 for $k=128$), we do not need to reduce the number of channels in the first transition layer.
However, in the later parts of the network the filter number is higher (\eg 640 at the second transition for $k=128$), so we use a slight reduction of 1.4 to keep the model size similar to a binary \architecture{ResNetE} of equal complexity.
On the other hand, we can replace the binary layer in this downsampling layer with a full-precision one (see \autoref{fig:transition-densenet}).
Since the full-precision convolution preserves more information, we can use reduction rates equal to (or even higher than) the reduction rate 2 of a full-precision \arch{DenseNet} for all downsampling layers.
These higher reduction rates also reduce the number of full-precision (and binary) weights and operations through the whole network, thus allowing us to reach a similar (or even lower) model size compared to the first approach.

Because of the previous reasons, we coupled the decision whether to use a binary or a full-precision downsampling convolution with the choice of reduction rate.
The two variants we compare in our experiments (see \autoref{sec:experiments}) are thus called \emph{full-precision downsampling with high reduction} (halve the number of channels in all transition layers) and \emph{binary downsampling with low reduction} (no reduction in the first transition, divide number of channels by 1.4 in the second and third transition).

\section{Experiments and Discussion}
\label{sec:experiments}







Following the structure of the previous section, we provide our experimental results to analyze our method with respect to different parameters and techniques.
We apply classification accuracy as the general measurement to evaluate the different architectures, methods etc.
For brevity, the term \emph{accuracy} always refers to the Top-1 accuracy, unless otherwise noted.
Also, differences in accuracies will be noted as $x$\%, but refer to percent point differences.
We use the MNIST~\cite{lecun-mnisthandwrittendigit-2010}, CIFAR-10~\cite{cifar10} and ImageNet~\cite{imagenet_cvpr09} datasets in terms of different levels of task complexity.
The experiments were performed on a work station with an Intel(R) Core(TM) i9-7900X CPU, 64 GB RAM and 4$\times$GeForce GTX1080Ti GPUs.
All models are trained with the Adam optimizer \cite{kingma2014adam} with an initial learning rate (alpha) of $10^{-2}$ for CIFAR-10 and $10^{-3}$ for ImageNet.
We trained our ImageNet models for 40 or 50 epochs, and multiply the learning rate by 0.1 at epochs 34 and 37, or epochs 40 and 45 respectively.
We use a Gaussian distribution to initialize the weights in the network according to the method proposed by Glorot and Bengio~\cite{glorot2010understanding}.

\begin{table}[]
\caption{Evaluation of our binary model performance on the MNIST and CIFAR-10 data sets compared to the results of Yang \etal \cite{HPI_xnor}.}
\begin{tabular}{|l|l|l|l|l|}
    \hline
                     & Architecture & \mrcell{Model\\size} & \mrcell{Accu-\\ racy} & \mrcell{Acc. \\ (\cite{HPI_xnor})} \\
    \hline
MNIST & LeNet  & 202KB & \textbf{99.0\%} & 97\% \\
    \hline
CIFAR-10 & ResNetE-18  & 1.39MB & \textbf{87.6\%} & 86\% \\ \cline{2-3} 
    \hline
CIFAR-10 & DenseNetE-21  & 1.49MB & \textbf{90.3\%} & - \\ \cline{2-3}    
    \hline 
\end{tabular} 
\label{tab:mnist-cifar-overview}
\end{table}

First, we show the results of a binary \emph{LeNet} for the MNIST dataset and a binary \architecture{ResNetE-18} and a \architecture{DenseNetE-21} in \autoref{tab:mnist-cifar-overview} compared to the approach of Yang \etal \cite{HPI_xnor}.
These results prove that our approach and implementation work on simple datasets, such as MNIST and CIFAR-10, and can reach favorable results compared to other work with the same approach and the same architecture.
Moreover, they reveal promising results with our proposed \architecture{DenseNetE} architecture, since the model size is increased by only 0.1MB for a 2.7\% increase in accuracy.

In the following sections, we first evaluate and discuss the influence of using scaling factors and the $\mathrm{approxsign}$ function in the backward pass of the activations for the \architecture{ResNetE} network.
Following this, we evaluate the impact of the amount of blocks for our proposed \architecture{DenseNet} architecture.
Then, the design choice of using binary or full-precision downsampling layers for \architecture{DenseNet} and \architecture{ResNetE} models is empirically verified.
Furthermore, we show how the bit-width of downsampling layers changes model performance.

\begin{table}[]
\caption{
    The influence of using scaling, a full-precision downsampling convolution, and the $\mathrm{approxsign}$ function on the CIFAR-10 dataset based on a \architecture{ResNetE-18}.
    Using $\mathrm{approxsign}$ instead of $\mathrm{sign}$ slightly boosts accuracy, but only if training a model with scaling factors. 
}
\begin{tabular}{|l|l|l|l|}
\hline
\lmrcell{Use \\ scaling \\ of \cite{Rastegari2016}} & \mrcell{Downsampl. \\ convolution} & \mrcell{Use \\ $\mathrm{approxsign}$ \\ of \cite{Liu_2018_ECCV}} & \mrcell{Accuracy \\ Top1/Top5} \\ \hline
\multirow{4}{*}{no}  & \multirow{2}{*}{binary}         & yes         & 84.9\%/99.3\%                   \\ \cline{3-4} 
                     &                                 & no          & 87.2\%/\textbf{99.5\%}          \\ \cline{2-4} 
                     & \multirow{2}{*}{full-precision} & yes         & 86.1\%/99.4\%                   \\ \cline{3-4} 
                     &                                 & no          & \textbf{87.6\%}/\textbf{99.5\%} \\ \hline
\multirow{4}{*}{yes} & \multirow{2}{*}{binary}         & yes         & 84.2\%/99.2\%                   \\ \cline{3-4} 
                     &                                 & no          & 83.6\%/99.2\%                   \\ \cline{2-4} 
                     & \multirow{2}{*}{full-precision} & yes         & 84.4\%/99.3\%                   \\ \cline{3-4} 
                     &                                 & no          & 84.7\%/99.2\%                   \\ \hline
\end{tabular}
\label{tab:cifar-scaling-downsampling-approxsign}
\end{table}

\begin{table}[]
\caption{
    The influence of using scaling, a full-precision downsampling convolution, and the $\mathrm{approxsign}$ function on the ImageNet dataset based on a \architecture{ResNetE-18}.
}
\begin{tabular}{|l|l|l|l|}
\hline
\lmrcell{Use \\ scaling \\ of \cite{Rastegari2016}} & \mrcell{Downsampl. \\ convolution} & \mrcell{Use \\ $\mathrm{approxsign}$ \\ of \cite{Liu_2018_ECCV}} & \mrcell{Accuracy \\ Top1/Top5} \\ \hline
\multirow{4}{*}{no}  & \multirow{2}{*}{binary}         & yes         & 54.3\%/77.6\%           \\ \cline{3-4}
                     &                                 & no          & 54.4\%/77.5\%           \\ \cline{2-4}
                     & \multirow{2}{*}{full-precision} & yes         & 56.6\%/\textbf{79.3\%}  \\ \cline{3-4}
                     &                                 & no          & \textbf{56.7\%}/79.2\%  \\ \hline
\multirow{4}{*}{yes} & \multirow{2}{*}{binary}         & yes         & 53.3\%/76.4\%           \\ \cline{3-4} 
                     &                                 & no          & 52.7\%/76.1\%           \\ \cline{2-4} 
                     & \multirow{2}{*}{full-precision} & yes         & 55.3\%/78.3\%           \\ \cline{3-4} 
                     &                                 & no          & 55.6\%/78.4\%           \\ \hline
\end{tabular}
\label{tab:imagenet-scaling-downsampling-approxsign}
\end{table}
\subsection{Scaling Methods}
\label{sec:results-scaling}

In this section, we discuss the influence of scaling factors (as proposed by Rastegari \etal \cite{Rastegari2016}) on the accuracy of our trained models based on the \architecture{ResNetE} architecture.
First, the results of our CIFAR-10 experiments verify our hypothesis, that applying scaling when training a model from scratch does not lead to better accuracy (see \autoref{tab:cifar-scaling-downsampling-approxsign}).
All models show a decrease of accuracy between 0.7\% and 3.6\% when applying scaling factors.
Secondly, we evaluated the influence of scaling for the ImageNet dataset (see \autoref{tab:imagenet-scaling-downsampling-approxsign}).
The result is similar, applying scaling reduces model accuracy ranging from 1.0\% to 1.4\%.
We conclude that the scaling is ineffective and suspect two arguments for this:
the model can not learn a useful scaling factor when training from scratch or the BatchNorm layers following each convolution layer absorb the effect of the scaling factors.
If the first reason applies this is a limitation of our approach of training from scratch, and might not apply to trainings based on fine-tuning a full-precision model.
If the latter reason applies it should neither increase nor decrease accuracy, which is what we can see for CIFAR-10 (but not for ImageNet) and might still help approaches which are based on fine-tuning.

\subsection{Backward Pass of the Sign Function}
\label{sec:results-approx}

In this section, we discuss the influence of the backward pass used for the $\mathrm{sign}$ function.
We compared the regular backward pass, called $\mathrm{sign}$, with the adapted backward pass, called $\mathrm{approxsign}$ (see \autoref{sec:implementation-details}).
First, the results of our CIFAR-10 experiments seem to depend on whether we use scaling or not.
If we use scaling, both functions perform similarly (see \autoref{tab:cifar-scaling-downsampling-approxsign}).
Without scaling the $\mathrm{approxsign}$ function leads to less accurate models on CIFAR-10.

In our experiments on ImageNet, the performance difference between the use of the functions is minimal (see \autoref{tab:imagenet-scaling-downsampling-approxsign}). 
Using one scaling method over the other gives no significant change in model accuracy with one exception: the usage of the sign function results in an accuracy increase of 0.6\% if we use scaling and no full-precision shortcut.
Therefore, we conclude that applying the $\mathrm{approxsign}$ function instead of the $\mathrm{sign}$ function seems to be specific to fine-tuning from full-precision models.

\begin{table}[]
\caption{
The accuracy of different binary \arch{DenseNet} models by successively splitting blocks evaluated on ImageNet.
As the number of connections increases, the model size (and number of binary operations) changes marginally, but the accuracy increases significantly.
}
\begin{tabular}{|l|l|l|l|}
\hline
\lmrcell{Blocks \\ (layers)} & \mrcell{Growth-\\ rate} & \mrcell{Model size \\ (binary)} & \mrcell{Accuracy \\ Top1/Top5}      \\ \hline
8 (13)   & 256         & 3.31 MB    & 50.2\%/73.7\% \\ \hline
16 (21)  & 128         & 3.39 MB    & 52.7\%/75.7\% \\ \hline
32 (37)  & 64          & 3.45 MB    &\textbf{54.3\%}/\textbf{77.3\%} \\ \hline
\end{tabular}
\label{tab:densenet-growth-rate-vs-layers}
\end{table}

\subsection{Splitting Layers of DenseNet}
\label{sec:results-split-densenet}

We tested our proposed architecture change by comparing \arch{DenseNet} models with varying growth rates and number of blocks (and thus layers).
The results show, that increasing the number of connections by adding more layers over simply increasing growth rate increases accuracy in an efficient way (see \autoref{tab:densenet-growth-rate-vs-layers}).
Doubling the number of blocks and halving the growth rate leads to an accuracy gain ranging from 1.4\% to 2.5\%.
However, it seems to have diminishing returns, and training of very deep binary \arch{DenseNet} becomes slow, since less of the calculations can be parallelized.
We note that during inference on low-powered devices this is less of a problem compared to training, since the total number of operations is similar between the models (and no additional memory is needed during inference for storing intermediate results, \eg the outputs of the sign function).
Therefore, we have not trained even more highly connected models, but highly suspect that this would increase accuracy even further.
The total model size slightly increases, since every second half of a split block has slightly more inputs compared to those of a double-sized normal block.
In conclusion, our technique of increasing number of connections is highly effective and size-efficient for a binary \arch{DenseNet}.

\begin{table}[]
\caption{
The difference of performance for different binary \arch{DenseNet} models when using different downsampling methods (see \autoref{sec:architectures-densenet}) evaluated on ImageNet.
}
\begin{tabular}{|l|l|l|l|}
\hline
\lmrcell{Blocks \\ (layers), \\ growth-rate} & \mrcell{Model \\ size \\ (binary)} & \mrcell{Downsampl. \\ convolution, \\ reduction} & \mrcell{Accuracy \\ Top1/Top5}      \\ \hline
\multirow{2}{*}{16 (21), 128}           & 3.39 MB    & binary, low   & 52.7\%/75.7\%                   \\ \cline{2-4}
                                        & 3.03 MB    & FP, high  & \textbf{55.9\%}/\textbf{78.5\%} \\ \hline
\multirow{2}{*}{32 (37), 64}            & 3.45 MB    & binary, low   & 54.3\%/77.3\%                   \\ \cline{2-4}
                                        & 3.08 MB    & FP, high  & \textbf{57.1\%}/\textbf{80.0\%} \\ \hline
\end{tabular}
\label{tab:densenet-downsampling}
\end{table}

\begin{table}[]
\caption{
Comparison on the ImageNet dataset \cite{imagenet_cvpr09} of our proposed network with binary downsampling branches (see \autoref{sec:architectures}) to ABC-Net \cite{lin2017towards}, which uses this design choice as well.
}
\begin{tabular}{|l|l|l|l|}
\hline
Model          & Our result (model size)                                  & ABC-Net \cite{lin2017towards} \\ \hline
ResNet-18      & 54.4\%/77.5\%                   (3.36 MB)                & 42.7\%/67.6\%                 \\ \hline
DenseNet       & 54.3\%/77.3\%                   (3.45 MB)                & -                             \\ \hline
ResNet-34      & 58.1\%/80.6\%                   (4.59 MB)                & -                             \\ \hline
\end{tabular}
\label{tab:imagenet-binary-downs}
\end{table}

\begin{table*}[]
\caption{
Comparison of our methods to state-of-the-art binary models on the ImageNet dataset \cite{imagenet_cvpr09}. 
All these methods use full-precision weights in the convolution layers of the downsampling branches (see \autoref{sec:architectures}).
}
\begin{center}
\begin{tabular}{|l|l|l|l|l|l|l|}
\hline
Model        & Our result                    (model size)             & BiReal-Net \cite{Liu_2018_ECCV} & TBN \cite{Wan_2018_ECCV}  & XNOR-Net \cite{Rastegari2016} & Full-precision \\ \hline
ResNet-18    & \textbf{56.9\%}/\textbf{79.7\%} (4.0 MB)                 & 56.4\%/79.5\%                   & 55.6\%/74.2\%             & 51.2\%/73.2\%                 & 69.3\%/89.2\%  \\ \hline
DenseNet     & \textbf{58.6\%}/\textbf{81.0\%} (3.99 MB)                & -                               & -                         & -                             & -              \\ \hline
ResNet-34    & 60.0\%/82.0\%                   (5.23 MB)                & \textbf{62.2\%}/\textbf{83.9\%} & 58.2\%/81.0\%       & -                             & 73.3\%/91.3\%  \\ \hline
\end{tabular}
\end{center}
\label{tab:imagenet-full-downs}
\end{table*}

\subsection{Downsampling Layers}
\label{sec:results-downsampling}

We evaluated the difference between using binary and full-precision downsampling layers for both \arch{ResNet} and \arch{DenseNet}.
First, we examine the results of \arch{ResNetE-18} on CIFAR-10.
Using full-precision downsampling over binary leads to an accuracy gain between 0.3\% and 2.3\% (see \autoref{tab:cifar-scaling-downsampling-approxsign}).
However, the model size also increases by 0.64 MB from 1.39 MB to 2.03 MB, which is is arguably too much for this minor increase of accuracy.
Our results on the \arch{ResNet} architecture show a significant difference on ImageNet (see \autoref{tab:imagenet-scaling-downsampling-approxsign}).
The accuracy increases by 2\% when using full-precision downsampling.
Similar to CIFAR-10, the model size increases by 0.64 MB, in this case from 3.36 MB to 4.0 MB.
The larger base model size makes the relative model size difference lower and provides a stronger argument for this trade-off.
We conclude that the increase in accuracy is significant, especially for ImageNet. 
However, in our opinion, it does not seem to be large enough to just neglect to acknowledge the significant increase in model size.

In the following we present our results of a binary \arch{DenseNet} when using a full-precision downsampling with high reduction over a binary downsampling with low reduction.
The results of a binary \arch{DenseNet-21} with growth rate 128 for CIFAR-10 result show an accuracy increase of 2.7\% from 87.6\% to 90.3\%.
The model size increases from 673 KB to 1.49 MB.
This is an arguably sharp increase in model size, but the model is still smaller than a comparable \arch{ResNet-18} with a much higher accuracy.
The results of two \arch{DenseNet} architectures (16 and 32 blocks combined with 128 and 64 growth rate respectively) for ImageNet show an increase of accuracy ranging from 2.8\% to 3.2\% (see \autoref{tab:densenet-downsampling}).
Further, because of the higher reduction rate, the model size decreases by 0.36 MB at the same time.
This shows a higher effectiveness and efficiency of using a full-precision downsampling layer for a \arch{DenseNet} compared to a \arch{ResNet}.

\subsection{Comparison to State-of-the-art Approaches}
\label{sec:results-state-of-the-art}

We evaluated our overall approach of training from scratch for a \arch{ResNetE-18}, a \arch{ResNetE-34} and our new architecture \arch{DenseNetE}.
Following our results on the influence of the downsampling convolution, we split the comparison between architectures with a full-precision and a binary downsampling convolution.

First, we would like to present the results for models with a binary downsampling convolution (see \autoref{tab:imagenet-binary-downs}).
In this case, we use the best \arch{DenseNetE-37} model with the highest number of connections as shown in our previous experiments (see \autoref{sec:results-split-densenet}).
It has a size comparable to that of a \arch{ResNet-18}.
We recognize that our training strategy of training from scratch leads to excellent results compared to the \arch{ABC-Net} \cite{lin2017towards} approach for both \arch{ResNets} with 18 and 34 layers.
The accuracy of our \arch{DenseNetE-37} is close to that of a \arch{ResNet} (with a difference of 0.2\%), but it does not improve accuracy.
This shows that the techniques applied to a \arch{ResNetE} and our \arch{DenseNetE-37} already successfully increase accuracy by a large margin, even without using full-precision downsampling layers.

Secondly, we also examine the results of models with full-precision downsampling layers (see \autoref{tab:imagenet-full-downs}).
We chose a growth rate of 160 and a reduction rate of 2.2 for a \arch{DenseNetE-21} to match the model size and complexity of a \arch{ResNetE-18} as closely as possible (3.99 MB and 4 MB respectively).
Our results show, that we can achieve results similar to a \arch{BiReal-Net} \cite{Liu_2018_ECCV} for 18 layers.
The accuracy is even slightly (0.5\%) higher, even though \arch{BiReal-Net} is trained with a more complex training strategy.
But, when we compare a \arch{ResNet-E} trained from scratch to a \arch{BiReal-Net} with 34 layers, we see that accuracy of our approach is 2\% lower in comparison.
Inspecting our training loss had us suspect, that this gap could be reduced by adapting our choice of optimizer, learning rates, and training for more epochs, but did not want to change this choice to keep our own results comparable.
Moreover, our proposed \arch{DenseNetE-21} model reaches 58.6\% (an overall improvement of 2.2\% over \arch{BiReal-Net-18}) with the same model size.
We conclude that accurate binary models can be successfully trained from scratch and do not necessarily need to use a fine-tuning strategy based on pretrained full-precision models.

\section{Conclusion}
\label{sec:conclusion}


In this paper, we presented our strategy to train binary neural networks from scratch.
We clearly separated the existing techniques to increase the number of connections of a binary \arch{ResNet} model and applied them to derive an accurate binary model based on a \arch{DenseNet} architecture.
Moreover, we showed the influence of these different techniques through comprehensive experiments and compared our approach to other state-of-the-art approaches.
We concluded that accurate binary models can be successfully trained from scratch and our proposed binary architecture even surpasses the state-of-the-art accuracy.
However, larger models can still benefit from complex fine-tuning strategies on pretrained full-precision models.

As future work, we would like to examine whether it is possible to better quantify the degree of importance of a layer in the network regarding the preservation of information.
Algorithmic approaches for this problem have already been proposed \cite{zhou2018adaptive}.
However, the theoretical knowledge would provide the basis to develop new architectures which benefit from this kind of model quantization.
Such a kind of novel architectures could help to reduce the accuracy gap between binary and full-precision layers.

{\small
\bibliographystyle{ieee}
\bibliography{egbib}
}

\end{document}